\documentclass{article}
\usepackage{spconf,amsmath,graphicx}
\usepackage{fancyhdr}
\usepackage{lipsum} % ダミーテキストのためのパッケージ

\title{E2GS: Event Enhanced Gaussian Splatting}
\name{Hiroyuki Deguchi*, Mana Masuda*, Takuya Nakabayashi, Hideo Saito}
\address{Keio University}
\usepackage{url}
\usepackage{caption}
\usepackage{amsmath}
\usepackage{xcolor}
\begin{document}
%\ninept
%
% \maketitle
%
\newcommand{\Ours}{Event Enhanced Gaussian Splatting}
\newcommand{\etal}{\textit{et al.}}
\newcommand{\red}[1]{\textcolor{red}{#1}}
\newcommand{\darkgraybox}[1]{\colorbox[rgb]{0.8,0.8,0.8}{#1}}
\twocolumn[{%
\renewcommand\twocolumn[1][]{#1}%
\maketitle
}]

\begin{abstract}

Event cameras, known for their high dynamic range, absence of motion blur, and low energy usage, have recently found a wide range of applications thanks to these attributes. In the past few years, the field of event-based 3D reconstruction saw remarkable progress, with the Neural Radiance Field (NeRF) based approach demonstrating photorealistic view synthesis results. However, the volume rendering paradigm of NeRF necessitates extensive training and rendering times. In this paper, we introduce Event Enhanced Gaussian Splatting (E2GS), a novel method that incorporates event data into Gaussian Splatting, which has recently made significant advances in the field of novel view synthesis. Our E2GS effectively utilizes both blurry images and event data, significantly improving image deblurring and producing high-quality novel view synthesis. Our comprehensive experiments on both synthetic and real-world datasets demonstrate our E2GS can generate visually appealing renderings while offering faster training and rendering speed (140 FPS).  Our code is available at \url{https://github.com/deguchihiroyuki/E2GS}.

% We propose Event Enhanced Gaussian Splatting, which is the first model to improve blurring for 3D Gaussian Splatting by utilizing the combination data of a bio-inspired event camera and a standard RGB camera.
\end{abstract}
%もう一個キーワード追加できる
\let\thefootnote\relax\footnotetext{*denots equal contribution}

\section{Introduction}
\label{sec:intro}
In the task of 3D scene reconstruction and novel view synthesis, we witnessed tremendous progress over the past few years. Especially, after the NeRF (Neural Radiance Field) \cite{NeRF} marked a significant milestone, leading to the active development of various neural rendering techniques for 3D scene reconstruction \cite{barron2021mip,pumarola2021d}. Among these, 3D Gaussian Splatting \cite{3Dgaussians} emerged as a simple yet computationally efficient method. It has gained recognition for its rapid training and rendering capabilities. However, these methods generally operate under ideal conditions and often struggle with motion blur, which can severely affect the quality of rendering.

Event cameras, inspired by biological vision systems, asynchronously capture changes in pixel brightness instead of recording absolute intensity at fixed frame rates as traditional frame-based RGB cameras do.
This unique approach offers several benefits over conventional cameras, including no-motion blur, high dynamic range, low power consumption, and lower latency. These advantages have spurred the development of various methods to address a range of computer vision challenges, such as optical-flow estimation \cite{gehrig2021raft}, and video interpolation \cite{lin2020learning}. To utilize these advantages, event cameras found their direction for development in 3D scene reconstruction tasks to handle high-speed camera movements or low lighting conditions which is hard for conditional RGB cameras \cite{e2nerf,eventnerf,klenk2022nerf}. While these methods showed photorealistic image rendering results compared to conditional RGB cameras in such conditions, they still require high computational complexity to train the whole network due to the ray-sampling strategy of the NeRF-based approach.

In this paper, we propose \Ours~(E2GS), the first approach that incorporates event data into Gaussian Splatting. By effectively incorporating the blurred RGB image and event data, our E2GS showed a visually appealing image deblurring and novel view synthesis result as shown in Fig.\ref{fig:teaser}. Our extensive experiments also showed our E2GS achieved better or competitive results while offering 60 times faster training and 3500 times faster rendering speed compared to $\textrm{E}^\text{2}$NeRF.

% もし場所余ったら
% Our contributions are summarized as follows:
% \begin{itemize}
%     \item 
% \end{itemize}

% \input{figs/overview}

% \twocolumn[\input{figs/teaser}]

\section{RELATED WORKS}
\label{sec:related}
\subsection{3D Scene Reconstruction}
3D scene reconstruction is one of the fundamental functionality of computer vision. Recent advancements in 3D scene reconstruction have gained more attention after the emergence of NeRF \cite{NeRF}. While several methods emerged to strengthen the NeRF-based approach \cite{barron2021mip,pumarola2021d}, there is one research direction to accelerate network training and image rendering speed \cite{muller2022instant}. Following this research interests, Kerbl \etal proposed 3D Gaussian Splatting \cite{3Dgaussians}, which eliminates the need for ray-sampling and instead uses Gaussians to present 3D space, which allows faster training and rendering.

\noindent\textbf{From Blurry Images}
We often observe blurriness in some parts or whole scenes when we casually take pictures. Various factors such as object motion, camera shake, and lens defocusing cause this blurriness. One conditional approach to deblurring images is to estimate the blur kernel or Point Spread Function (PSF) and deconvolve the image. Some works have been proposed to deblur images with the training of the 3D
\twocolumn[% \begin{figure}
%   \centering
%   \includegraphics[width=1.0\linewidth]{figs/teaser.pdf}
%   \caption{When we take as input blurry images of a scene from multiple views, the rendering results of original 3D Gaussian Splatting \cite{3Dgaussians} are also severely blurred. In contrast, our E2GS achieves sharper scene rendering by utilizing event data.}
% \end{figure}
\begin{center}
  \centering
  \includegraphics[width=0.95\linewidth]{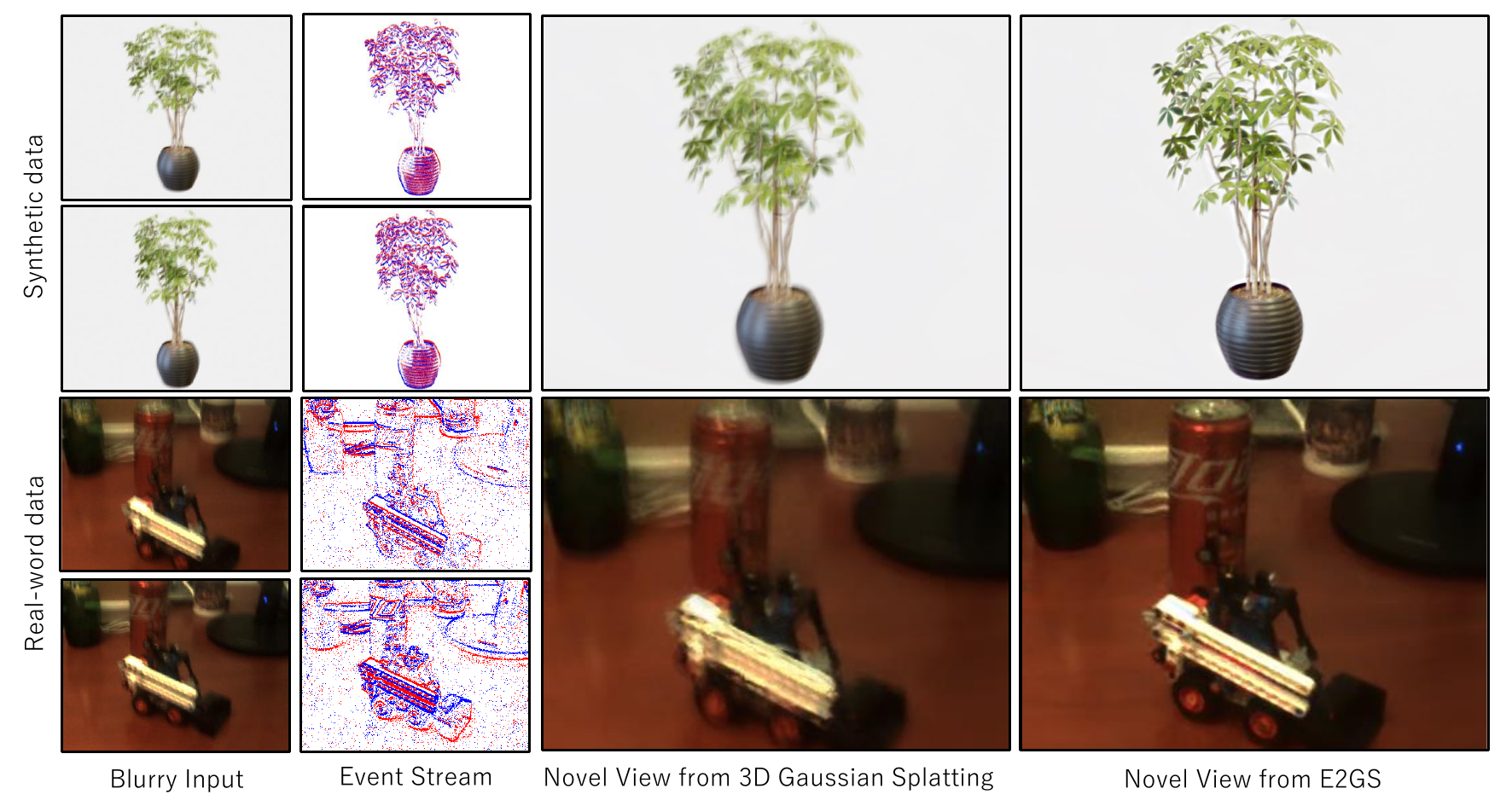}
  \captionof{figure}[t]{When we take as input blurry images of a scene from multiple views, the rendering results of original 3D Gaussian Splatting \cite{3Dgaussians} are also severely blurred. In contrast, our E2GS achieves sharper scene rendering by utilizing event data.}
  \label{fig:teaser}
\end{center}]
\noindent
scene reconstruction framework. Deblur-NeRF\cite{deblurNeRF} is a pioneering work that employs an additional MLP
to estimate per-pixel blur kernel. Lee \etal~\cite{lee2024deblurring} proposed to use additional MLP to manipulate the covariance of each Gaussian to model blurriness.

\subsection{Event-based 3D Scene Reconstruction}
Event cameras, also known as dynamic vision sensors (DVS) \cite{EventCamera}, asynchronously capture pixel brightness changes, drawing inspiration from biological vision systems. This unique recording framework effectively addresses the issue of information loss between frames, a common problem in frame-based RGB cameras. Event cameras offer several benefits, including no motion blur, high dynamic range, low power consumption, and reduced latency. Due to these advantages, they have shown remarkable results in various tasks like optical flow estimation \cite{opticalflow}, depth estimation \cite{depth}, and feature detection and tracking \cite{feature}. Recently, Ev-NeRF \cite{evnerf} and EventNeRF \cite{eventnerf} have managed to train NeRF models solely using the event data. However, these methods experience noticeable artifacts and chromatic aberration, and they also exhibit limited generalization ability in pose estimation for neural representation learning. Meanwhile, $\textrm{E}^\text{2}$NeRF \cite{e2nerf} has successfully trained a sharper NeRF by utilizing both blurry RGB images and corresponding event data. Despite this advancement, it still suffers from prolonged training and rendering times due to the ray-sampling-based NeRF rendering strategy.

\begin{figure*}
  \centering
  \includegraphics[width=1.0\linewidth]{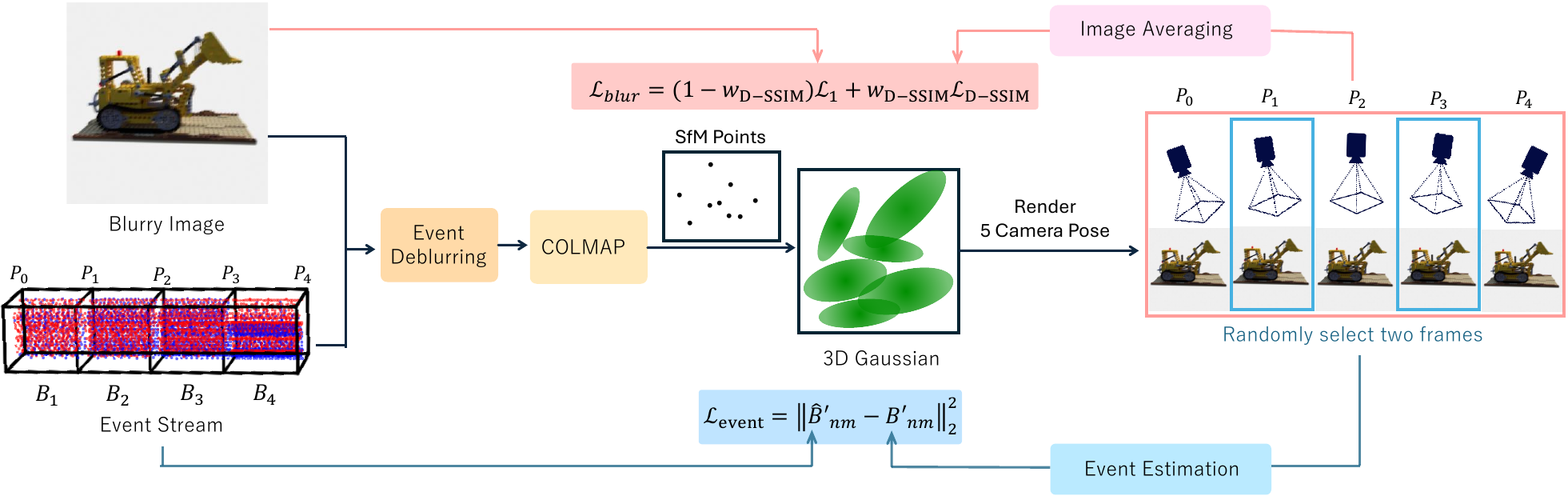}
  \caption{The overview of the \Ours.}
  \label{fig:overview}
\end{figure*}

\section{METHOD}
\label{sec:method}

The overview of our method is illustrated in Fig. \ref{fig:overview}. The input of our method is a set of blurry images and event stream of a static scene. In our E2GS framework, we first perform preprocessing using the correspondence between event data and blurred images. Then, we use two types of loss functions to train the Gaussian Splatting considering the blur.

\subsection{Preliminary}
\noindent\textbf{3D Gaussian Splatting.}
To represent a volumetric scene and render it, we adopt methods from 3D Gaussian Splatting, which proposes differentiable rasterization. The Gaussians are defined by a full 3D covariance matrix $\mathbf{\Sigma}$ defined in world space \cite{ewa}:
\begin{equation} \label{Def of Gaussians}
  G(\mathbf{x}) = e^{-\frac{1}{2}\mathbf{x}^{T}\mathbf{\Sigma}^{-1}\mathbf{x}}.
\end{equation}
 To render the novel views, the covariance matrix in the camera coordinates of the novel view can be obtained as:
% To project these 3D gaussians to 2D for rendering, the projected covarience matrix ${\Sigma}^{\prime}$ in camera coordinates is given as:
\begin{equation} \label{Def of Sigma dash}
  {\mathbf{\Sigma}}^{\prime} = \mathbf{J}\mathbf{W}{\mathbf{\Sigma}}\mathbf{W}^T\mathbf{J}^T.
\end{equation}
where $\mathbf{J}$ is the Jacobian of the affine approximation of the projective transformation and $\mathbf{W}$ is the viewing transform matrix. To directly optimize the $\mathbf{\Sigma}$, it is expressed as:
\begin{equation} \label{Def of sigma}
  \mathbf{\Sigma} = \mathbf{R}\mathbf{S}\mathbf{S}^T\mathbf{R}^T,
\end{equation}
where $\mathbf{S}$ is the scaling matrix and $\mathbf{R}$ is the rotation matrix.
% レンダリングまでの流 れ
% GS引用してる論文のGS説明の流れみてく
% \subsection{Event Data}

\noindent\textbf{Event Data.}
Event cameras asynchronously report an event $e(x,y,\tau,p)$ when they detect the brightness changes of pixel $(x,y)$  exceeds the threshold $C$ at time $\tau$. Instead of reporting the actual intensity value $L(x,y,\tau)$, they report intensity change direction $p$ which is defined as follows;
\begin{equation}\label{Def of event}
    p(x,y,\tau) = 
    \begin{cases}
    +1 & \text{if}~l(x,y,\tau) - l(x,y,\tau') > C \\
    -1 & \text{if}~l(x,y,\tau) - l(x,y,\tau') < -C
    \end{cases},
\end{equation}
where $l(x,y,\tau) = \log(L(x,y,\tau))$ and $\tau'$ represents the timestamp of the last observed event at pixel $(x, y)$.

% \subsection{Image Deblurring}
% Given a set of blurry images, we prepare $N$ timestamps $\{t_i\}_{i=1}^N$ which divides the event stream into $N-1$ event bins $\{B_i\}_{i=1}^{N-1}$ for a more accurate estimate of the intensity change during the exposure time:
% \begin{equation}
%     B_{i}= \{e_j(x_j,y_j,\tau_j,p_j)\}_{j=1}^{N_e^i} ({t_{i}<\tau_i\leq{t_{i+1}}}),
% \end{equation}
% where $N_e^i$ indicates the number of events in $i$-th event bin.
% To estimate $N$ camera poses at each time $t_i$, we use Event-based Double Integral (EDI) \cite{EDI}. The EDI model assumes that the blurred image is the average of the multiple sharp images during the exposure time and that a sharp image at a certain time is expressed by adding events. From this assumption, the image $I_i$ at the moment $t_i$ can be expressed as follows:
% \begin{equation}\label{sharp}
%   I_{i+1} = I_{i} + \sum_{j=1}^{N_e^{i}}\exp(Cp_j).
% \end{equation}
% The blurry image $I_\textit{blur}$ can be expressed as the average of the images at each timestamp since we set each timestamp to equally divide the exposure time:
% \begin{equation}
%     I_\textit{blur} = \frac{1}{N}\sum_{i=1}^{N} I_i.
% \end{equation}

% \subsection{Camera Pose and Initial Point Cloud Estimation}
% To estimate the initial 3D Gaussian coordinate and camera pose, we feed all deblurred image sets $\{I_i\}_{i=1}^N$ of each blurry image to COLMAP Structure-from-Motion package \cite{COLMAP}. Without the image deblurring stage, the COLMAP often fails as reported in \cite{e2nerf}.

\subsection{Preprocessing}
To utilize a framework for high temporal resolution event data, it is necessary to prepare the initial point cloud for Gaussian splatting and N equally spaced camera poses during the exposure time of each viewpoint. The specific steps are detailed below.

\noindent\textbf{Image Deblurring.}
Given a set of blurred images and event stream corresponding to the exposure time of each image, we prepare $N$ timestamps $\{t_i\}_{i=1}^N$ which divides the event stream equally into $N-1$ event bins $\{B_i\}_{i=1}^{N-1}$ for a more accurate estimate of the intensity change during the exposure time:
\begin{equation}
    B_{i}= \{e_j(x_j,y_j,\tau_j,p_j)\}_{j=1}^{N_e^i} ({t_{i}<\tau_i\leq{t_{i+1}}}),
\end{equation}
where $N_e^i$ indicates the number of events in $i$-th event bin.
To estimate $N$ camera poses at each time $t_i$, we use Event-based Double Integral (EDI) \cite{EDI}. The EDI model assumes that the blurred image is the average of multiple sharp images during the exposure time. Furthermore, based on the relationship between the event data and the change in brightness described in Eq. \ref{Def of event}, it is assumed that a sharp image at a certain time can be represented by adding events. From this assumption, the image $I_i$ at the moment $t_i$ can be expressed as follows:
\begin{equation}\label{sharp}
  I_{i+1} = I_{i}\sum_{j=1}^{N_e^{i}}\exp(Cp_j).
\end{equation}
The blurry image $I_\textit{blur}$ can be expressed as the average of the images at each timestamp since we set each timestamp to equally divide the exposure time:
\begin{equation}
    I_\textit{blur} = \frac{1}{N}\sum_{i=1}^{N} I_i.
\end{equation}

\noindent\textbf{Camera Pose and Initial Point Cloud Estimation.}
To estimate the initial 3D Gaussian coordinate and camera pose, we feed all deblurred image sets $\{I_i\}_{i=1}^N$ of each blurry image to COLMAP Structure-from-Motion package \cite{COLMAP}. Without the image deblurring, the COLMAP often fails as reported in \cite{e2nerf}
\subsection{Loss function}
To learn the scene from blurred images, we use two types of losses: Image Rendering Loss and Event Rendering Loss.

\noindent\textbf{Image Rendering Loss.}
To adapt 3D Gaussian Splatting taking blurry images as input, we introduce image rendering loss. With $N$ camera poses ${\{P_i\}}^N_{i=1}$ of each view, we render $N$ rendered images ${\{\hat{I}_i\}}^N_{i=1}$. Since we set each timestamp to equally divide the exposure time, we can estimate the blurry image by taking the average of the $N$ rendered images:
\begin{equation}
\label{I_blur}
  \hat{I}_\textit{blur} = \frac{1}{N}\sum_{i=1}^N \hat{I}_i,
\end{equation}
% Comparing $\hat{I}_{blur}$ and supervision $I_{blur}$, the loss function is L1 combined with a D-SSIM term, with reference to 3D Gaussian Splatting \cite{3Dgaussians}:
The loss function is L1 loss combined with a D-SSIM loss which compares $\hat{I}_\textit{blur}$ and supervision $I_\textit{blur}$.
The final image rendering loss $\mathcal{L}_\textit{blur}$ is written as follows using a weight loss parameter $w_\text{D-SSIM}$:
\begin{equation}
\label{L_blur}
  \mathcal{L}_\textit{blur} = (1-w_\text{D-SSIM})\mathcal{L}_1+w_\text{D-SSIM}\mathcal{L}_\text{D-SSIM}.
\end{equation}

\noindent\textbf{Event Rendering Loss.}
While image rendering loss simply averages $N$ images to simulate a blurred frame, it does not take into account the temporal process of blurring. To utilize event information to supervise the continuous blurring process with high temporal resolution, we employed event loss. Given estimated images from $N$ poses, we first randomly select the two frames $\{I_n, I_m\}$ ($n<m$) from ${\{\hat{I}_i\}}^N_{i=0}$ and convert them into grayscale intensity images $L_n$ and $L_m$. We take the difference of the two intensity values in the log domain and divide it by the threshold $C$ for each pixel $(x, y)$ to estimate the number of events between two frames:
\begin{equation}\label{estimate event}
  \hat{B}'_{nm}=
  \begin{cases}
    \lfloor{\frac{\log(L_m)-\log(L_n)}{C}}\rfloor & \text{if $L_n (x,y)<L_m(x,y)$} \\
    \lceil{\frac{\log(L_m)-\log(L_n)}{C}}\rceil       & \text{if $L_n(x,y)\geq L_m(x,y)$}
  \end{cases}.
\end{equation}
We use the mean squared error to evaluate the error between estimated event bin image $\hat{B}'_{nm}$ and ground truth event bin image ${B}'_{nm}$, storing an actual number of events for each pixel. Note that there are cancelations of the positive and negative events in the GT event bin image since our model assumes the monotonic intensity change between the timesteps:
\begin{equation} 
\label{L_event}
  \mathcal{L}_\textit{event} = \|\hat{B}'_{nm}-{B}'_{nm}\|^2_2.
\end{equation}

Finally, we combine two loss function $\mathcal{L}_{blur}$ and $\mathcal{L}_\textit{event}$ by using a weight parameter $w_\textit{event}$ to obtain the following loss 

\begin{equation} 
\label{eq:loss}
\mathcal{L} = \mathcal{L}_{blur} + w_\textit{event}\mathcal{L}_\textit{event},
\end{equation}

\section{EXPERIMENTS}
\label{sec:experiments}

% synthetic real_dataそれぞれについての評価
% novel_viewについても行う
% 定量評価，定性評価
% train_time, rendering_time

\subsection{Experimental Setup}
We evaluated our E2GS on two different tasks: Image deblurring and novel view synthesis. For the image deblurring task, we evaluate the rendering results from the perspective of the blurry image set. for the novel view synthesis task, we evaluate the rendering results from the perspective not used in the blurry image set.

\noindent\textbf{Implementation Details.}
Our code is based on 3D Gaussian Splatting \cite{3Dgaussians}. We train each scene with 30k iterations on a single NVIDIA RTX A5000 GPU. For all data, we set $w_\text{D-SSIM}=0.2$, $w_\textit{event}=0.005$, and $N = 5$. We set the different thresholds for positive and negative events to estimate the event bin image $C_\textit{pos}=0.2, C_\textit{neg} = 0.3$. The rest of the parameters follow the 3D Gaussian Splatting default values.

\noindent\textbf{Comparison Methods.}
To evaluate the effectiveness of utilizing event data to solve the image deblurring task and novel view synthesis task, we compared our model with normal Gaussian Splatting (GS) \cite{3Dgaussians}, which takes blurry images as input. Note that we obtained the initial point cloud and camera poses by using deblurred images of EDI same as our methods since COLMAP often fails when we use blurry images. The other comparison method is $\textrm{E}^\text{2}$NeRF \cite{e2nerf}, which is a state-of-the-art method that solves the image deblurring and the novel view synthesis tasks by utilizing a NeRF-based approach. We also report ``GS w/ $\mathcal{L}_\textit{blur}$'' result which only uses blur Loss $\mathcal{L}_\textrm{blur}$ to evaluate the effectiveness of the event loss $\mathcal{L}_\textrm{event}$.

% For comparison, we first choose ``GS'', which simply inputs blurry images to normal Gaussian Splatting \cite{3Dgaussians}. (Initial point cloud and camera poses are obtained by EDI deblurred images because COLMAP fails when we input blurry images.)
% We also compare our method to $\textrm{E}^\text{2}$NeRF \cite{e2nerf}, which is a state-of-the-art NeRF method for learning sharp scenes from blurry images. In addition, ``GS w/ $\mathcal{L}_\textit{blur}$'' only uses Blur Loss (Sec.\ref{blur_loss}), while "Event Enhanced GS" both uses Blur Loss and Event Loss (Sec.\ref{event_loss}).
% Except for GS with deblurred images, we input blurry images.

\noindent\textbf{Evaluation Metrics.}
To quantitatively evaluate the quality of the rendered image we employed three extensively recognized metrics to evaluate image quality for the synthetic dataset: Peak Signal-to-Noise Ratio (PSNR), Structural Similarity Index Measure (SSIM), and the Learned Perceptual Image Patch Similarity (LPIPS) \cite{LPIPS}. Since the real-world data does not contains ground truth sharp images, we use Blind/Referenceless Image Spatial Quality Evaluator (BRISQUE) \cite{brisque}, which evaluates the naturalness of the image without any references based on the distribution of the brightness.

\subsection{Datasets}
To evaluate the effectiveness of our method, we use $\textrm{E}^\text{2}$NeRF \cite{e2nerf} dataset.

% \begin{table*}[ht]
% \centering
% \caption{Quantitative evaluation of our method on the image deblurring. The results in the table are the averages of the six synthetic scenes from NeRF \cite{NeRF}. We use \textbf{bold} to mark the best data.}
% \begin{tabular}{c||ccccc}
% \hline
% Image Deblurring & GS\cite{3Dgaussians} & GS w/ deblurred image & $\textrm{E}^\text{2}$NeRF \cite{e2nerf} & GS w/ $\mathcal{L}_\textit{blur}$ & E2GS (Ours) \\ \hline
% PSNR$\uparrow$ & 22.92 & 29.26 & 29.77 & 30.20   & \textbf{30.83}   \\
% SSIM$\uparrow$ & 0.886 & 0.954 & \textbf{0.960} & 0.951   & 0.957    \\
% LPIPS$\downarrow$& 0.105 & \textbf{0.052} & 0.073 & 0.064   & 0.059     \\ \hline
% \end{tabular}
% \label{tab:deblur}
% \end{table*}

\begin{table}[tb]
\centering
\caption{Quantitative evaluation of our method on the image deblurring. The results in the table are the averages of the six synthetic scenes from NeRF \cite{NeRF}.}
\scalebox{0.85}[0.85]{
\begin{tabular}{c||cccc}
\hline
Image Deblur & GS & $\textrm{E}^\text{2}$NeRF & GS w/ $\mathcal{L}_\textit{blur}$ & E2GS (Ours) \\ \hline
PSNR$\uparrow$ & 22.92 & 29.77 & 30.20   & \darkgraybox{30.84}   \\
SSIM$\uparrow$ & 0.886 & \darkgraybox{0.960} & 0.951   & 0.957    \\
LPIPS$\downarrow$& 0.105 & 0.073 & 0.064   & \darkgraybox{0.059}     \\ \hline
\end{tabular}}
\label{tab:deblur}
\end{table}

\begin{table}[tb]
\centering
\caption{Quantitative evaluation of our method on the novel view synthesis. The results in the table are the averages of the six synthetic scenes from NeRF \cite{NeRF}.}
\scalebox{0.85}[0.85]{
\begin{tabular}{c||cccc}
\hline
View Synthesis & GS & E${}^\text{2}$NeRF & GS w/ $\mathcal{L}_\textit{blur}$ & E2GS (Ours)\\ \hline
PSNR$\uparrow$ & 22.15 & \darkgraybox{29.56} & 28.33   & 28.89   \\
SSIM$\uparrow$ & 0.878 & \darkgraybox{0.962} & 0.943   & 0.949    \\
LPIPS$\downarrow$& 0.113 & 0.073 & 0.071 & \darkgraybox{0.069}     \\ \hline
\end{tabular}}
\label{tab:novel_view}
\end{table}

% \begin{table*}[ht]
% \centering
% \caption{Quantitative evaluation of our method on the novel view synthesis. The results in the table are the averages of the six synthetic scenes from NeRF \cite{NeRF}. We use \textbf{bold} to mark the best data.}
% \begin{tabular}{c||ccccc}
% \hline
% Novel View Synthesis & GS \cite{3Dgaussians} &GS w/ deblurred image & E${}^\text{2}$NeRF \cite{e2nerf} & GS w/ $\mathcal{L}_\textit{blur}$ & E2GS (Ours)\\ \hline
% PSNR$\uparrow$ & 22.15 &29.56 & \textbf{29.56} & 28.33   & 28.83   \\
% SSIM$\uparrow$ & 0.878 & 0.962 & \textbf{0.962} & 0.943   & 0.948    \\
% LPIPS$\downarrow$& 0.113 & \textbf{0.054} & 0.073 & 0.071   & 0.069     \\ \hline
% \end{tabular}
% \label{tab:novel_view}
% \end{table*}

% \begin{table}[tb]
% \centering
% \caption{Quantitative evaluation of our method on the novel view synthesis. The results in the table are the averages of the six synthetic scenes from NeRF \cite{NeRF}.}
% \scalebox{0.85}[0.85]{
% \begin{tabular}{c||cccc}
% \hline
% View Synthesis & GS & E${}^\text{2}$NeRF & GS w/ $\mathcal{L}_\textit{blur}$ & E2GS (Ours)\\ \hline
% PSNR$\uparrow$ & 22.15 & \darkgraybox{29.56} & 28.33   & 29.22   \\
% SSIM$\uparrow$ & 0.878 & \darkgraybox{0.962} & 0.943   & 0.950    \\
% LPIPS$\downarrow$& 0.113 & 0.073 & 0.071 & \darkgraybox{0.068}     \\ \hline
% \end{tabular}}
% \label{tab:novel_view}
% \end{table}
\begin{table}[t]
\centering
\caption{Quantitative evaluation of the image deblurring task. Showing the BRISQUE results of five scenes from $\textrm{E}^\text{2}$NeRF \cite{e2nerf} and the average of the five scenes.}
\scalebox{0.77}[0.77]{
\begin{tabular}{c||ccccc|c}
\hline
Image Deblur& letter & lego  & camera & toys  & plant & Avg. \\ \hline
GS   & 40.68  & 39.52 & 21.76  & 43.66 & 38.26 & 36.78  \\
$\textrm{E}^\text{2}$NeRF & 44.33  & \darkgraybox{34.09} & 28.89  & 43.41 & 32.23 & 36.59   \\
E2GS (Ours) & \darkgraybox{37.62}  & 35.2  & \darkgraybox{19.93}  & \darkgraybox{38.87} & \darkgraybox{30.87} & \darkgraybox{32.50}  \\ \hline
\end{tabular}}
\label{tab:real_d}
\end{table}

\begin{table}[h]
\centering
\caption{Quantitative evaluation of the novel view synthesis task. Showing the BRISQUE results of five scenes from $\textrm{E}^\text{2}$NeRF \cite{e2nerf} and the average of the five scenes.}
\scalebox{0.76}[0.76]{
\begin{tabular}{c||ccccc|c}
\hline
  View Synthesis   & letter & lego  & camera & toys  & plant & Avg. \\ \hline
GS   & 40.83  & 39.02 & 22.01  & 44.28 & 39.25 & 37.08  \\
$\textrm{E}^\text{2}$NeRF  & 44.19  & \darkgraybox{34.23} & 28.77  & 43.42 & \darkgraybox{32.03} & 36.53  \\
E2GS (Ours) & \darkgraybox{37.10}   & 35.64 & \darkgraybox{19.90}   & \darkgraybox{38.7}4 & 32.49 & \darkgraybox{32.77}  \\ \hline
\end{tabular}}
\label{tab:real_novel}
\end{table}

% \begin{table*}[]
% \centering
% \caption{Quantitative evaluation of our method by BRISQUE$\downarrow$. The results are the averages of five scenes on all views.}
% \begin{tabular}{c||ccccc|c}
% \hline
% & letter & lego  & camera & toys  & plant & Average \\ \hline
% GS   & 40.80  & 39.12 & 21.96  & 44.16 & 39.05 & 37.02   \\
% $\textrm{E}^\text{2}$NeRF & 44.22  & 34.20 & 28.79  & 43.42 & 32.07 & 36.54   \\
% E2GS(Ours) & 37.20  & 35.55 & 19.91  & 38.77 & 32.17 & 32.72   \\ \hline
% \end{tabular}
% \label{tab:real}
% \end{table*}
\begin{table}[!h]
\centering
\caption{Training and rendering time evaluation.}
\begin{tabular}{c||cc}
\hline
 & E${}^\text{2}$NeRF & E2GS (Ours) \\ \hline
% Training time(min) & 2903  & \textbf{50 min}               \\
Training time & 2 days  & \textbf{50 min}               \\
Rendering (FPS)  & 0.04 & \textbf{140}   \\ \hline       
\end{tabular}
\label{tab:time}
\end{table}

\noindent{\textbf{Synthetic data:}} Synthetic set contains six synthetic scenes (chair, ficus, hotdog, lego, materials, and mic), and it uses the Camera Shakify plugin in Blender to simulate camera shake. The event data are simulated by V2E\cite{v2e}. Each scene has 100 views of blurry images estimated by 17 different camera poses from the Camera Shakify plugin, its corresponding event data, and camera poses.
% , 17 camera poses during the camera shaking and blurred image simulated by reproducing isp processing. 
% Each scene has 100 views of 800×800 blurry images and the corresponding event data and camera poses. For each view, we select 5 poses between 4 equal time intervals during the blurring process.

\noindent{\textbf{Real-world data:}} Real-world set contains five challenging scenes (letter, lego, camera, plant, and toys) captured by DAVIS346 color event camera \cite{DAVIS}. Each scene has 30 views of blurry images and the corresponding event data. 

% It has five challenging scenes (letter, lego, camera, plant, and toys) captured by hand. It contains event data and blurry images obtained from DAVIS346 color event camera \cite{DAVIS}. Each scene has 30 views of 346×260 blurry images and the corresponding event data and camera poses.\par
% In addition, in order to get the initial point cloud, we first use EDI \cite{EDI} model to get 5 deblurred images between 4 equal time per view and input those images to COLMAP to get 5 poses. Because when we use blurry images, reconstruction of 3D points and camera pose estimation often fails.

\subsection{Quantitative Evaluation}
\noindent\textbf{Synthetic data:} Tab. \ref{tab:deblur} shows the result of the image deblurring and Tab. \ref{tab:novel_view} shows the result of the novel view synthesis. 
% For comparison, we first choose ``GS'', which simply inputs blurry images to normal Gaussian Splatting \cite{3Dgaussians}. (Initial point cloud and camera poses are obtained by EDI deblurred images because COLMAP fails when we input blurry images.)
% For comparison, we first choose ``GS with deblurred image'', which input images deblurred by the model of Nakabayashi \etal~\cite{nakabayashi} to the normal Gaussian Splatting. 
% We also compare our method to $\textrm{E}^\text{2}$NeRF \cite{e2nerf}, which is a state-of-the-art NeRF method for learning sharp scenes from blurry images. In addition, ``GS w/ $\mathcal{L}_\textit{blur}$'' only uses Blur Loss (Sec.\ref{blur_loss}), while "Event Enhanced GS" both uses Blur Loss and Event Loss (Sec.\ref{event_loss}). 
% Except for GS with deblurred images, we input blurry images. 
% We employed three extensively recognized metrics to evaluate image quality: Peak Signal-to-Noise Ratio (PSNR), Structural Similarity Index Measure (SSIM), and the Learned Perceptual Image Patch Similarity (LPIPS) \cite{LPIPS}. 
Tab. \ref{tab:deblur} shows the result on the image deblurring task, E2GS achieves better or comparable results with $\textrm{E}^\text{2}$NeRF. Tab. \ref{tab:novel_view} shows the result on the novel view synthesis task, E2GS achieves better or comparable results with $\textrm{E}^\text{2}$NeRF. On both tasks, E2GS outperforms both GS and GS w/ $\mathcal{L}_\textit{blur}$ in all three metrics, which shows the effectiveness of utilizing events and event loss to render novel views from blurry image frames.

% We use PSNR, SSIM, LPIPS \cite{LPIPS} for evaluation.
% 表の結果

\noindent\textbf{Real-world data:} Tab. \ref{tab:real_d} and 
Tab. \ref{tab:real_novel} shows the quantitative result of real-world data for the image deblurring task and the novel view synthesis task respectively. E2GS outperformed other comparable methods for both tasks.

%(blur viewとnovel viewでどれもあんまりBRISQUE結果違わなかったので，表1つにしました)
\subsection{Qualitative Evaluation}
\begin{figure}[t]
  \centering
  \includegraphics[width=1.00\linewidth]{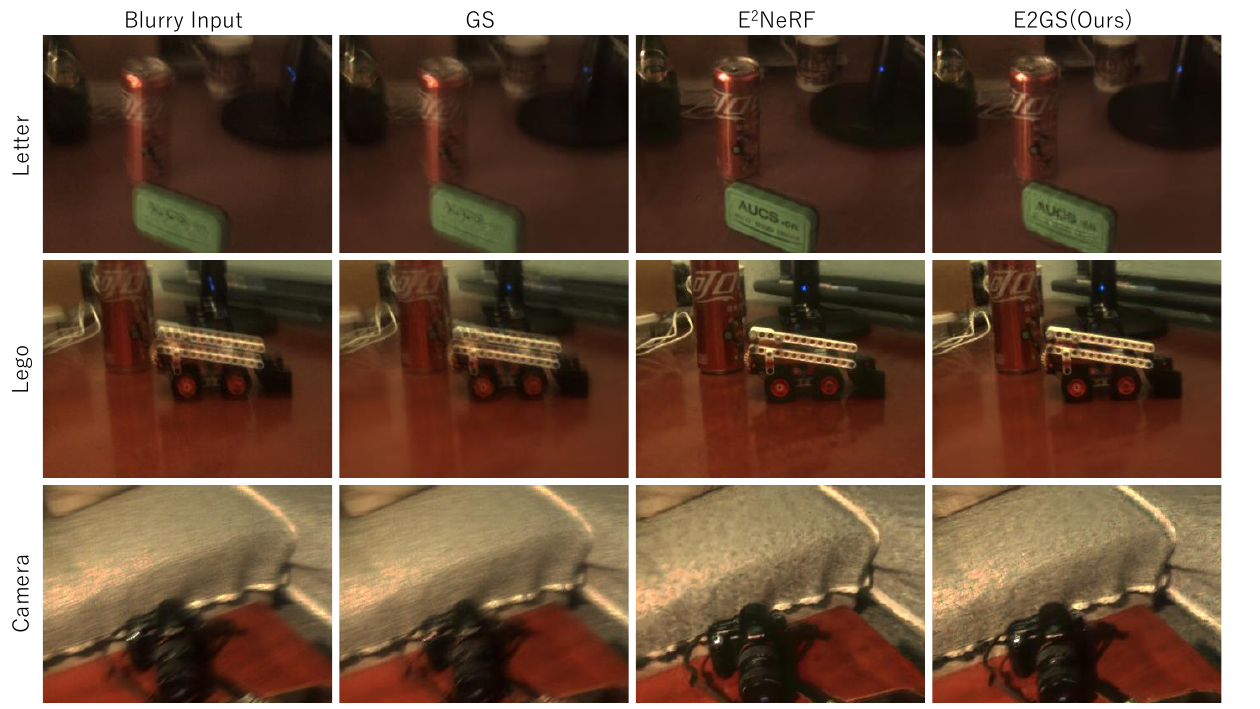}
  \caption{Qualiative comparison of the image deblurring task on the real-world dataset.}
  \label{fig:realworld_blur}
\end{figure}

\begin{figure}[!h]
  \centering
  \includegraphics[width=0.95\linewidth]{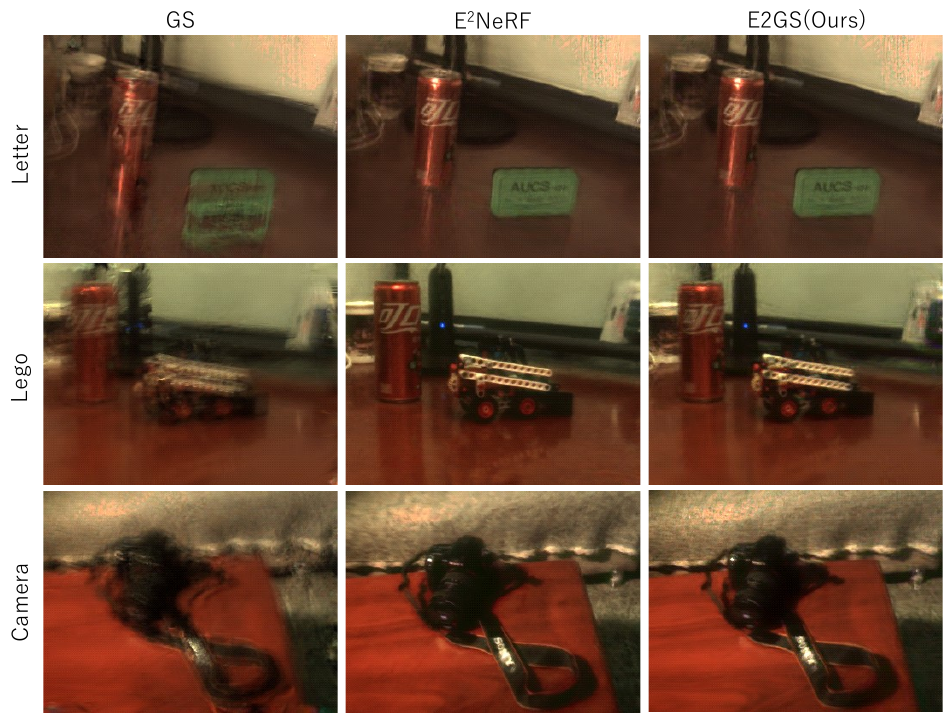}
  \caption{Qualiative comparison of the novel view synthesis task on the real-world dataset.}
  \label{fig:realworld_novel}
\end{figure}

\begin{figure*}[h]
  \centering
  \includegraphics[width=1.0\linewidth]{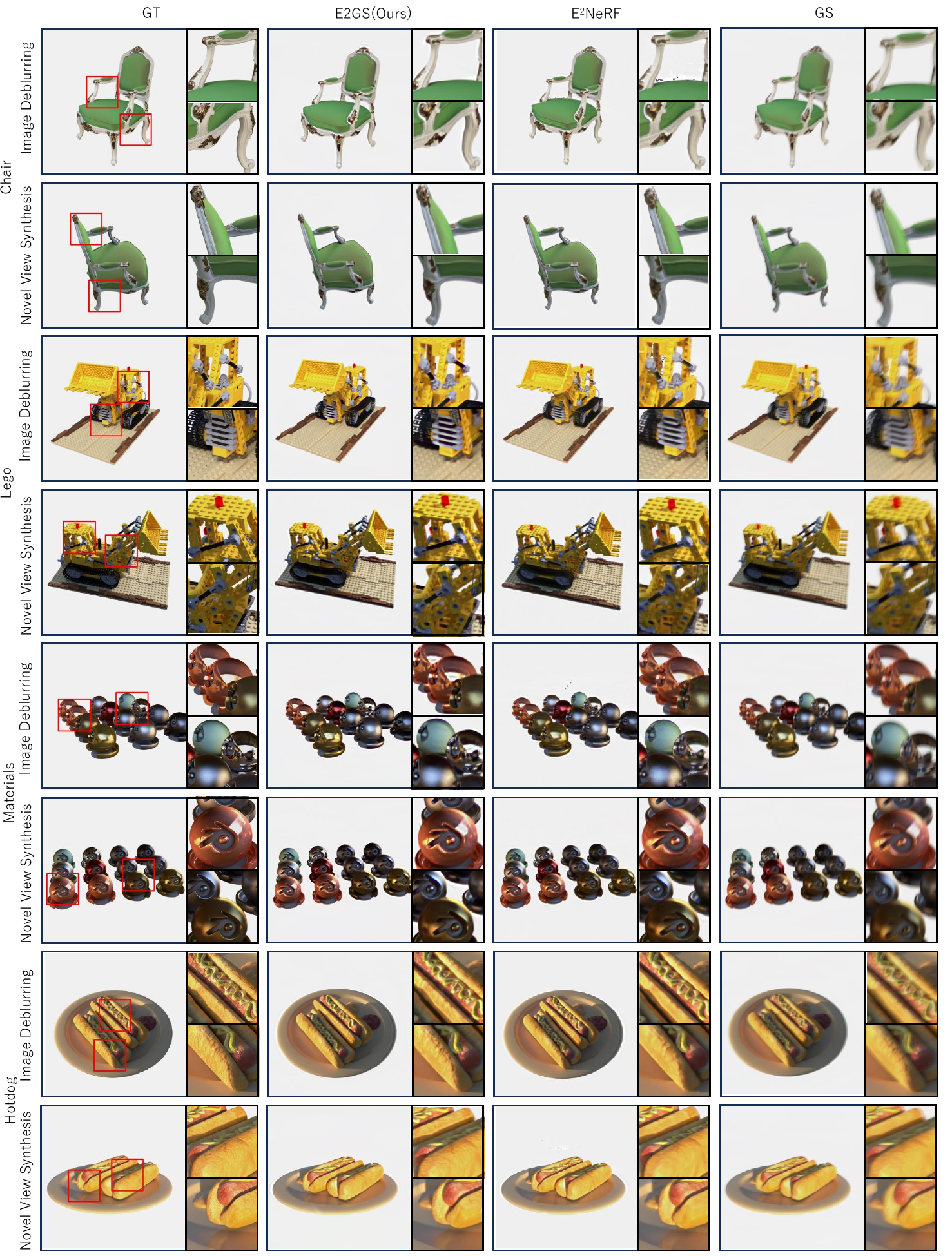}
  \caption{Qualiative comparison on the synthetic dataset. Refer to the red box to see the detailed reconstruction quality. Zoom in for the best view.}
  \label{fig:synthetic}
\end{figure*}

\noindent \textbf{Synthetic data: }
We report the rendering result of synthetic data of our E2GS and two baseline methods in Fig. \ref{fig:synthetic}. GS produces reasonable rendering results from their blurry RGB inputs. $\textrm{E}^\text{2}$NeRF is achieved to reconstruct the sharp image by utilizing the event data, but they fail to reconstruct the details of the scenes, e.g. small parts and reflection of the surface.
% In contrast, we can find that E2GS renders the scenes as sharp as $\textrm{E}^\text{2}$NeRF, which is consistent with the results of quantitative analysis. We can also find that our E2GS can reconstruct realistically the details of the scenes such as small parts of the lego, the reflection of the materials and the surface of the hotdog bun.
\noindent \textbf{Real-world data: }
Fig. \ref{fig:realworld_blur} and Fig. \ref{fig:realworld_novel} show the rendering result of the real-world dataset on the image deblurring task and the novel view synthesis task respectively. Our E2GS achieves to render sharp images for both tasks. 
% that default GS reconstructs blurry scenes, but our E2GS renders sharp images and has as good visual performance as $\textrm{E}^\text{2}$NeRF on both tasks.
% E2NeRFの方が字のところとかくっきりできてはいる，，

\subsection{Training Time and Rendering Speed}
Tab. \ref{tab:time} shows training time and rendering FPS of $\textrm{E}^\text{2}$NeRF and our E2GS. For this evaluation, we use the synthetic dataset with 800 $\times$ 800 resolution as input. Thanks to the rasterizing-based image rendering framework, our E2GS drastically reduces both training time and rendering time compared to $\textrm{E}^\text{2}$NeRF. More specifically, our E2GS reduced the training time to 1/60, and the rendering speed to 1/3500 compared to $\textrm{E}^\text{2}$NeRF.

\section{CONCLUSION}
\label{sec:conclusion}

In this paper, we propose \Ours~(E2GS), the novel framework that effectively utilizes event data into Gaussian Splatting to reconstruct sharp scenes from blurry RGB frames. Comprehensive experiments using the synthetic dataset and the real-world dataset demonstrate that our E2GS achieves visually appealing rendering quality and significantly faster training and rendering speed (140 FPS) compared to previous state-of-the-art methods. Future research directions include addressing dynamic scenes with fast-moving subjects, e.g. sports scenes, which are challenging to handle only by using RGB frame cameras.

% We believe that our method can be applied in the real-time rendering of a complex scene. Finally, our method can only be applied to a static scene. Therefore, We will apply our method to a dynamic scene in the future.
% 最後に応用領域みたいなことここに書く？？Future direction, 静的なのにしか対応してないからみたいなこととか

\vspace{12pt}
\noindent\textbf{Acknowledgment}
This work was partly supported by JST SPRING, Grant Number JPMJSP2123, and JSPS KAKENHI Grant Number JP23H03422.

\vfill\pagebreak
\clearpage
% \section{REFERENCES}
% \label{sec:refs}

\bibliographystyle{IEEEbib}
\bibliography{strings,refs,masuda}

\end{document}